\documentclass[pdflatex,sn-mathphys-num]{sn-jnl}

\usepackage{graphicx}%
\usepackage{multirow}%
\usepackage{amsmath,amssymb,amsfonts}%
\usepackage{amsthm}%
\usepackage{makecell}
\usepackage{mathrsfs}%
\usepackage[title]{appendix}%
\usepackage{xcolor}%
\usepackage{textcomp}%
\usepackage{manyfoot}%
\usepackage{booktabs}%
\usepackage{listings}%
\usepackage{makecell}

\usepackage[linesnumbered,ruled,vlined]{algorithm2e}
\usepackage{xcolor}
\newcommand{\answer}[1]{\textcolor{blue}{#1}}

\theoremstyle{thmstyleone}%
\theoremstyle{thmstyletwo}%

\theoremstyle{thmstylethree}%

\begin{document}

\title[Article Title]{TypiCore: A Hybrid Active Query Strategy for Class-Incremental Learning on Time Series}


\author*[1]{\fnm{Gábor} \sur{Szűcs}}\email{szucs@tmit.bme.hu}
\author[1]{\fnm{Sámuel} \sur{Jacsev}}\email{jacsev.samuel@edu.bme.hu}
\author[1]{\fnm{Marcell} \sur{Németh}}\email{nemethm@tmit.bme.hu}

\author[2]{\fnm{Davide} \sur{Dalle Pezze}}\email{davide.dallepezze@unipd.it}

\author[2]{\fnm{Gian Antonio} \sur{Susto}}\email{gianantonio.susto@unipd.it}

\affil[1]{\orgdiv{Department of Telecommunications and Artificial Intelligence},
\orgname{Budapest University of Technology and Economics},
\orgaddress{\city{Budapest}, \country{Hungary}}}

\affil[2]{\orgdiv{Department of Information Engineering},
\orgname{University of Padua},
\orgaddress{\city{Padua}, \country{Italy}}}


\abstract{Time series data play a pivotal role across numerous domains, including healthcare and manufacturing. In real-world environments, models must cope with distribution shifts over time, a challenge commonly addressed through Continual Learning (CL) techniques. However, existing CL methods face a critical limitation: real-world data streams are rarely fully labeled, making annotation cost a major practical constraint. 
This paper investigates Active Class-Incremental Learning (ACIL) for multivariate time series, where a model must sequentially learn new classes while selectively querying labels under a fixed annotation budget.
We present a systematic evaluation of a wide range of query strategies combined with multiple rehearsal-based approaches, assessing their impact on plasticity, stability, and label efficiency across four benchmark datasets. 
Our analysis reveals the limitations of uncertainty-based and distribution-aware methods in achieving strong performance under constrained labeling budgets. To address these shortcomings, we propose TypiCore, a novel hybrid query strategy that alternates between typicality-based and diversity-based sample selection across active learning cycles, enabling the construction of memory buffers that are both representative and diverse. 
Evaluated on the TSCIL benchmark, TypiCore delivers significant improvements over all baselines and matches or surpasses fully supervised continual learning performance on multiple datasets while requiring a fraction of the available labels.}




\maketitle

\section{Introduction}
\label{sec:intro}

The deployment of machine learning systems in dynamic real-world environments increasingly demands models capable of adapting to continuously evolving data distributions. 
This challenge is especially pronounced in the time series domain, where applications such as industrial fault detection, healthcare monitoring, and human activity recognition operate in inherently non-stationary settings. In these scenarios, new classes of events or behaviors emerge over time, a new fault type in a production line, a previously unseen activity pattern, or a novel clinical condition, and models must incorporate this new knowledge without catastrophically overwriting what was previously learned. 

How to adapt to new incoming data while retaining the previous knowledge is known as Continual Learning (CL).
Among its various formulations, Class-Incremental Learning (CIL) represents one of the most challenging and practically relevant settings: the model must sequentially learn new classes as they arise, while remaining capable of classifying all previously seen classes at inference time, and without access to task identifiers \cite{van2022three}. 

Despite substantial progress in CL in recent years \cite{sulaiman2025online}, there are two critical issues that are unexplored.
The first is that classic CL scenarios assume incoming data streams are fully labeled, but in practice, this assumption rarely holds. 
Data acquisition is often automatic and inexpensive, but annotation requires domain expertise, sustained manual effort, and considerable time and financial resources.
The second gap concerns the scope of evaluation. The vast majority of continual learning research has been conducted in the computer vision \cite{pezze2025continual}, text domain \cite{yang2023continual} (e.g. named entity recognition \cite{liu2025class}) and, more recently, natural language processing domains \cite{carta2026adapting}, leaving time series data comparatively underexplored despite its prevalence in high-stakes applications.
One of the few exceptions is the work \cite{qiao2024class}, which introduced the Time Series Class-Incremental Learning (TSCIL) benchmark to standardize evaluation across a diverse set of time series datasets. However, the benchmark operates under the standard fully-labeled assumption, leaving the label-scarce setting entirely unaddressed.

Active Learning (AL) offers a principled solution to the labeling bottleneck by strategically selecting the most informative samples for annotation, maximizing model performance under a fixed labeling budget \cite{settles2009active} or smallest number of labeled samples \cite{badine2026salient}. In isolation, both CIL and AL are well-established research areas with rich literature and mature methodologies. Yet their intersection, Active Class-Incremental Learning (ACIL), remains remarkably underexplored, with few works focused on image classification \cite{bhattacharya2026acil}. 
However, time series classification introduces additional complexity beyond standard CIL benchmarks: temporal dependencies, multivariate channel interactions, inter-subject variability, and noisy acquisition conditions all compound the challenge of learning incrementally from limited labeled data.

The first contribution of this work is a comprehensive and systematic evaluation of active learning query strategies in rehearsal-based approaches for the CIL scenario applied to multivariate time series classification.
The core of the work involves designing and systematically evaluating a wide range of active learning methodologies, including uncertainty-based, distribution-aware, and hybrid strategies.

We then analyze the effectiveness of these strategies in dynamic environments using the TSCIL benchmark.
The first insight is that, contrary to the expectation, complex and optimal methods like ASER show that in data-scarce active learning scenarios, the simpler ER mechanism often yields superior stability.
The uncertainty-based methods prove consistently detrimental, frequently degrading performance below random selection.
In contrast, distribution-aware methods like Core-Set excel at geometric coverage, maximizing exposure to the underlying data distribution while hybrid approaches like TypiClust excels at representativeness by selecting typical high-density samples, thereby promoting stable memory retention.
However, neither approach alone achieves a robust balance between plasticity and stability across the benchmark.

Therefore, as the third contribution, motivated by this analysis, we introduce TypiCore, a novel hybrid active query strategy that alternates between typicality-based and diversity-based selection across active learning cycles.
TypiCore constructs memory buffers that are simultaneously robust against forgetting and expressive enough to capture the complexity of the underlying data distribution.

Through experiments on the TSCIL benchmark, we show that TypiCore consistently outperforms all baseline methods with statistically significant gains across multiple datasets. Furthermore, despite operating under a limited labeling budget, it matches or exceeds the performance of fully supervised continual learning approaches on several benchmarks. The source code of our TypiCore method with the complete framework is publicly available on a repository for reproducibility.

The remainder of this paper is organized as follows. Section~\ref{sec:rw} reviews the relevant literature on continual learning, covering the main scenarios and approaches.
 Section~\ref{sec:active_learning} surveys the active learning strategies, from uncertainty-based methods to diversity and typicality-based approaches.
 Section~\ref{sec:methodology} formalizes the proposed Active Class-Incremental Learning framework,  and introduces TypiCore, our novel hybrid query strategy. 
 Section~\ref{sec:exp_setting} details the experimental setup, including the benchmark datasets, model architecture, training configuration, and evaluation protocol. 
 Section~\ref{sec:results} presents and analyzes the empirical results across all query strategies and datasets. 
 Finally, Section~\ref{sec:conclusion} concludes the paper, summarizing the key findings and outlining directions for future work.

\begin{figure}[!thbp]
    \centering
    \includegraphics[width=0.99\textwidth]{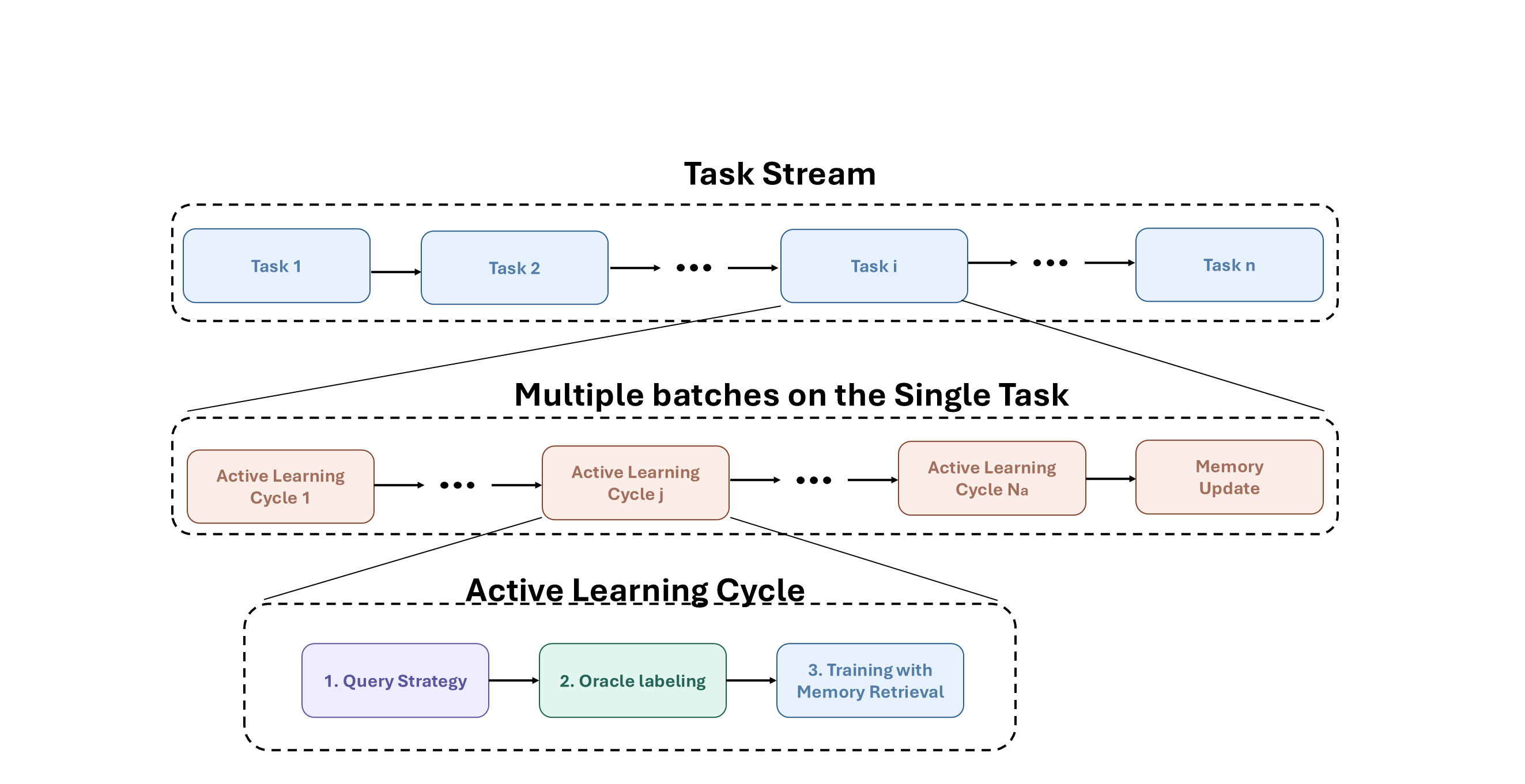}
    \caption{\textbf{ACIL Framework}. Scheme to integrate Active Learning strategies in the Class Incremental Learning scenario. Each task is composed of multiple active learning cycles and each one is composed of three phases: i) Query Strategy, ii) Oracle labeling, iii) Training.  }
    \label{fig:scheme}
\end{figure}

\section{Related Work}
\label{sec:rw}

Traditional machine learning paradigms assume independent and identically distributed
samples drawn from a stationary distribution, but real-world applications often encounter
evolving data streams where new tasks emerge sequentially.

The ability to learn sequentially from non-stationary data streams without forgetting previously acquired knowledge has emerged as a central challenge in modern machine learning and is known as Continual Learning (CL).
The core problem is catastrophic forgetting, where a neural network trained sequentially on new tasks tends to lose performance on prior information.
This phenomenon occurs because the neural network abruptly overwrites previously learned representations when exposed to new training data. Therefore, CL focuses on developing techniques to mitigate forgetting during network updates and find an optimal balance between stability (preserving old knowledge) and plasticity (acquiring new knowledge) \cite{yamauchi2023quick}.

\vspace{\baselineskip}

Literature on CL usually adopt the taxonomy proposed in \cite{van2022three} introducing three CL scenarios, (i)
Task-Incremental Learning (TIL), where task identity is available at inference time, (ii) Domain-Incremental Learning (DIL), where the class set remains fixed but the input distribution shifts, and (iii) Class-Incremental Learning (CIL), where new classes arrive sequentially and the model must classify among all classes seen so far without task identity. 
Of these, CIL represents the most challenging and practically relevant setting, as it most closely mirrors real-world conditions.

While continual learning research has historically been concentrated in computer vision, and more recently in natural language processing driven by the rise of large language models, other data modalities have received comparatively little attention. Time series data, despite its prevalence in real-world applications \cite{song2026semi, szucs2025distribution} such as industrial monitoring, healthcare, and activity recognition, remains much less explored in the continual learning literature. To address this gap, the Time Series Class-Incremental Learning (TSCIL) benchmark \cite{qiao2024class} was developed, which systematically evaluates the performance of established continual learning methods across four curated time series datasets under the class-incremental learning scenario.

\vspace{\baselineskip}

To mitigate catastrophic forgetting, the literature has converged on three broad families of approaches: regularization-based, architecture-based and rehearsal-based \cite{wickramasinghe2023continual}.
\textbf{Regularization-based methods} impose constraints on parameters or activations to protect knowledge critical to previous tasks. 
Some well-known methods belonging to this category are Elastic Weight Consolidation (EWC) \cite{kirkpatrick2017overcoming}, which introduced a penalty on parameters important for prior tasks, and Learning without Forgetting (LwF) \cite{li2017learning}, which instead applies knowledge distillation \cite{hinton2015distilling}, constraining the network's output activations on new data to remain consistent with those of the previously trained model. 
However, these methods tend to degrade under long task sequences and have been shown to perform poorly in the CIL scenario.

\textbf{Architecture-based methods} address forgetting by allocating distinct model components to different tasks, ensuring that learning a new task does not interfere with the parameters storing knowledge from previous ones \cite{,fernando2017pathnet}.
Representative approaches include Progressive Neural Networks \cite{rusu2016progressive}, which grow the architecture with each new task by adding lateral connections to frozen columns, and PackNet \cite{mallya2018packnet}, which iteratively masks and prunes subnetworks within a fixed-capacity model.
However, these methods have two fundamental limitations: they usually do not scale well with the number of tasks, and task identity is typically required at inference time to route inputs to the correct subnetwork,  an assumption not allowed in the CIL scenario.

\textbf{Rehearsal-based methods} have emerged as the most practically effective family of approaches for CIL \cite{chaudhry2019tiny, buzzega2020dark}. The core principle, introduced by Experience Replay (ER) \cite{chaudhry2019tiny}, is simple yet powerful. A small subset of samples from previous tasks is stored in a fixed-size episodic memory buffer. These samples are revisited together with the new samples when visiting a new task.  In this way, the model can retain the accumulated knowledge by periodically revisiting previous data while remaining plastic enough to incorporate new classes.

Despite its simplicity, ER has proven a remarkably competitive baseline across diverse benchmarks, motivating a rich line of follow-up work focused on improving memory management strategies.
While some methods consider hybrid approaches that combine replay with other strategies \cite{rebuffi2017icarl,lopez2017gradient,buzzega2020dark}, a parallel line of work has focused specifically on improving memory retrieval  \cite{aljundi2019online,shim2021online}.
For example,  ASER leverages an efficient k-nearest-neighbor approximation of Shapley values to quantify each sample's marginal contribution to classification performance \cite{shim2021online} and achieves state-of-the-art performance on the TSCIL benchmark \cite{qiao2024class}. 

\vspace{\baselineskip}

Despite the significant advances, a fundamental limitation persists across the entire continual learning literature: the assumption that incoming data streams are fully labeled. In practice, however, deploying a continual learning system requires continuous human intervention to annotate new data, a process that is costly, time-consuming, and difficult to scale. This bottleneck severely constrains the real-world applicability of existing CIL methods, regardless of their methodological complexity. 
Therefore, in this paper, we directly address this limitation by incorporating active learning capabilities in the TSCIL benchmark.
We analyze the effect of different query strategies on the performance of rehearsal-based methods, with a particular focus on ER and ASER.
Finally, we propose TypiCore, a novel hybrid query strategy that alternates between typicality-based and diversity-based sample selection across active learning cycles, enabling the construction of memory buffers that are both representative and diverse.

\section{Active Learning Approaches}
\label{sec:active_learning}

\subsection{Introduction}
In classification problems, obtaining labels can be expensive, often requiring expert knowledge.
Therefore, given a fixed labeling budget, the specific subset of data chosen for labeling can critically impact the resulting model’s performance \cite{kumari2026deep}.

Active Learning is the process of strategically selecting the most informative samples from an unlabeled pool for labeling, thereby maximizing learning performance with minimal labeling effort \cite{settles2009active, cherman2019multi}.
Active learning systems implement this selection using query strategies in an iterative
process. The model is trained on an initial set of labeled samples, resulting in a model
state $\theta$. Based on this model state, the active learning strategy selects a new batch
of samples from the unlabeled pool, from which the model is expected to gain the most
information. These strategies generally fall into two main categories: \textbf{uncertainty-based}
and \textbf{distribution-aware}.

\subsection{Uncertainty-Based Methods}

Uncertainty-based methods select samples about which the model, in its current state $\theta$,
is most uncertain, based on the assumption that uncertain samples are likely to be the most
informative for improving model performance.

These methods rely on the model’s softmax output probabilities computed using the model state from the previous step, $\theta_{a-1}$. In the following, $p_{\theta_{a-1}}(y_k \mid x)$ denotes the predicted probability for class $k$ given the input $x$. We define $p_{\theta_{a-1}}(y_{\max1} \mid x)$ and $p_{\theta_{a-1}}(y_{\max2} \mid x)$ as the probabilities of the most (maximum predicted probability) and second-most likely classes, respectively.

\textbf{Entropy-Based Uncertainty}
Entropy measures the disorder or uncertainty in the model’s predicted probability distribution \cite{tharwat2023survey}. For a classification task with $K$ classes, the uncertainty is defined as follows, where samples with high entropy are considered uncertain.
\begin{equation}
U_{\text{entropy}}(x) = -\sum_{k=1}^{K} p_{\theta_{a-1}}(y_k \mid x) \log p_{\theta_{a-1}}(y_k \mid x)
\end{equation}

\textbf{Margin-Based Uncertainty}
This strategy focuses on the difference between the top two predicted classes \cite{tharwat2023survey}. The next equation presents the formula, where samples with small margins are considered uncertain. 
\begin{equation}
U_{\text{margin}}(x) = p_{\theta_{a-1}}(y_{\max1} \mid x) - p_{\theta_{a-1}}(y_{\max2} \mid x)
\end{equation}

\textbf{Least Confidence}
This strategy selects samples where the model is least confident in its most probable prediction \cite{tharwat2023survey}. Samples with low confidence are considered uncertain, where the confidence is calculated based on the next equation.
\begin{equation}
U_{\text{LC}}(x) = 1 - p_{\theta_{a-1}}(y_{\max1} \mid x)
\end{equation}

\subsection{Distribution-aware Methods}

Distribution-aware methods select samples that best represent the overall data distribution in the feature space, ensuring good coverage and preventing the model from focusing on narrow regions \cite{settles2009active}.
Distribution-aware methods aim to select samples that capture the variety and span the underlying structure of the overall data distribution in the feature space. 

\textbf{Core-Set} \cite{sener2017active} (as diversity-based method) employs a greedy $k$-center approximation algorithm to find a subset that effectively covers the data by choosing centers that minimize the maximum distance from any data point to its nearest center. The algorithm operates in the feature space extracted from the model's embedding layer using a furthest-first traversal strategy.

Given labeled samples $L$ and unlabeled candidates $U$, the algorithm maintains for each unlabeled sample the distance to its nearest labeled neighbor. At each iteration, it selects the unlabeled sample with the maximum such distance, effectively choosing the sample that is furthest from all currently labeled points. After adding this sample to $L$, the distances are updated incrementally by comparing each unlabeled sample's current minimum distance with its distance to the newly selected point. This process repeats until $n$ samples are selected.

Formally, at each iteration, the algorithm selects:
\begin{equation}
x^* = \arg\max_{x \in U} \min_{x' \in L} \|f(x) - f(x')\|_2
\end{equation}
where $f(\cdot)$ denotes the feature embedding. This greedy approach ensures maximum coverage of the feature space and provides a 2-approximation to the optimal $k$-center problem.

Some methods combine different objectives, such as uncertainty, diversity \cite{he2014active, giouroukis2025dual}.
However, although combining uncertainty and diversity can enhance performance, these two criteria alone may be insufficient in low-budget scenarios. 

Therefore, \textbf{TypiClust} \cite{hacohen2022active} (typicality-based method) introduces a third dimension, representativeness, which proves equally critical: rather than selecting samples that are merely uncertain or geometrically spread, a robust query strategy should prioritize samples that are typical of their class distribution.
\textit{Representativeness} describes how typical a sample is of its class, or equivalently, how similar it is to other samples in the same region of the feature space.
Building on this principle, TypiClust \cite{hacohen2022active} combines clustering with typicality scoring to systematically identify and select the most representative samples from each region of the data distribution. 

Typicality measures a sample’s density in feature space, quantified as the inverse of the average Euclidean distance to its $K$ nearest neighbors:

\begin{equation}
\label{eq:typicality}
\text{Typicality}(x) = \left( \frac{1}{K} \sum_{x_i \in \text{$K$-NN}(x)} \|f(x) - f(x_i)\|_2 \right)^{-1}
\end{equation}
where $f(x)$ is the feature vector, $\text{$K$-NN}(x)$ denotes the $K$ nearest neighbors of $x$, and $\|f(x) - f(x_i)\|_2$ is the Euclidean distance between $x$ and neighbor $x_i$ in the feature space. The method proceeds through three steps:
\begin{enumerate}
    \item \textbf{Representation learning:} Learn a feature space from the unlabeled pool $\mathcal{U}_0$ using self-supervised learning.
    \item \textbf{Clustering for diversity:} At iteration, partition the data into $k$ clusters, where $k$ is the sum of the batch size and the number of labeled examples. This ensures that at least the number of clusters equal to the batch size do not contain labeled examples.
    \item \textbf{Querying typical examples:} Select the most typical example (Eq. \ref{eq:typicality}) from each of the largest clusters (corresponding to the batch size $b$) without labeled examples.
\end{enumerate}

\begin{algorithm}[H]
\label{alg:er}
\DontPrintSemicolon
\caption{Generic ER-based method for Task $t$}
\KwIn{Memory $\mathcal{M}$; Parameters $\theta_{t-1}$, Dataset $\mathcal{D}_t$; Task $t$}
\BlankLine

Initialize parameters: $\theta' \leftarrow \theta_{t-1}$\;

\While{not converged}{
    \For{$b \sim \mathcal{D}_t$}{
        \If{$t = 1$}{
            $\theta' \leftarrow \text{SGD}(b, \theta')$ \tcp*{first task, empty memory}
        }
        \Else{
            $B_{\mathcal{M}} \leftarrow \text{MemoryRetrieval}(\mathcal{M}, b, \theta')$\;
            $\theta' \leftarrow \text{SGD}(B_{\mathcal{M}} \cup b, \theta')$ \tcp*{train on combined batch}
        }
    }
}
$\theta_t \leftarrow \theta'$ \tcp*{converged, set as optimal for task t}
\BlankLine

\For{$b \sim \mathcal{D}_t$}{
    $\mathcal{M} \leftarrow \text{MemoryUpdate}(\mathcal{M}, b, \theta')$ \tcp*{memory update}
}

\Return $\theta_t, \mathcal{M}$
\end{algorithm}

\section{Methodology: TypiCore in ACIL Framework}
\label{sec:methodology}

\subsection{Replay Approach}

Among the approaches proposed to address catastrophic forgetting, replay-based methods have emerged as one of the most effective and practical.
The core idea is to maintain a fixed-size memory buffer $\mathcal{M}$ containing a small subset of samples from previously learned tasks. 
During training on a new task, previously stored exemplars are retrieved from the memory buffer and interleaved with the current task's data, exposing the model simultaneously to old and new class distributions and thereby counteracting the forgetting.

Formally, the sequential learning scenario consists of $T$ incremental tasks. Each task $t \in \{1, \ldots, T\}$ introduces a disjoint subset of new classes $\mathcal{C}_t$, such that $\mathcal{C}_i \cap \mathcal{C}_j = \emptyset$ for all $i \neq j$. The set of all classes encountered up to and including task $t$ is defined as $\mathcal{C}_{\leq t} = \bigcup_{i=1}^{t} \mathcal{C}_i$. At the onset of each task $t$, a dataset $\mathcal{D}_t = \{\mathbf{x}_i^t\}_{i=1}^{N_t}$ of $N_t$ samples becomes available. Each sample $\mathbf{x} \in \mathbb{R}^{d \times \tau}$ is a multivariate time series with $d$ channels and $\tau$ time steps.

Note: even though continuous streams may contain overlapping classes, our ACIL framework assigns incremental tasks completely disjoint class subset; since this allows for a clear measurement of forgetting during evaluation, and it was also defined exactly this way in previous articles \cite{bhattacharya2026acil, szHucs2025active}.

The learner's objective is to incrementally train a classifier $f_\theta : \mathbb{R}^{d \times \tau} \rightarrow \mathcal{C}_{\leq t}$ that, after completing task $t$, correctly classifies instances from the full joint label space $\mathcal{C}_{\leq t}$. 
Furthermore, we operate under the class-incremental learning protocol, in which no task identifier is provided at inference time and the model must discriminate among all previously seen classes without access to task identifiers.

In our framework, we used two variants of the Replay Approach, (i) Adversarial Shapley Value Experience Replay (ASER)~\cite{shim2021online} and (ii) Experience Replay (ER)~\cite{chaudhry2019tiny}, which are described below.

\vspace{\baselineskip}

\subsection{Experience Replay}

Experience Replay (ER)~\cite{chaudhry2019tiny} is the most famous replay-based method and consists of two components.
\\
 \textbf{1. Memory Retrieval.}  During training on task $t$, each mini-batch $B_\ell$ sampled from the current dataset $\mathcal{D}_t$ is augmented with a batch $B_m$ retrieved uniformly at random from the memory buffer $\mathcal{M}$. The model is then updated on the combined batch $B_m \cup B_\ell$, exposing it simultaneously to new and old class distributions. 
The model  parameters $\theta$ are updated by minimizing the combined loss:
\begin{equation}
\label{eq:combined_loss}
    \mathcal{L}(\theta_t) = \lambda \mathcal{L}_{\text{CE}}(\theta_t; B_m) + (1-\lambda) \mathcal{L}_{\text{CE}}(\theta_t; B_\ell) 
\end{equation}
where $\mathcal{L}_{\text{CE}}$ denotes the cross-entropy loss, and  $\lambda$ is the weight parameter of the instances in the memory buffer. 
Our methodology combines two objectives with a unified optimization criterion as shown in Eq. \ref{eq:combined_loss}, causing the balance between both objectives to be sufficiently controlled. 
\\
\textbf{2. Memory Update.} After processing each task, the buffer $\mathcal{M}$ is updated via reservoir sampling~\cite{vitter1985random}, which guarantees that every observed sample has an equal probability of being retained regardless of when it was encountered. This ensures that $\mathcal{M}$ remains a uniformly random subset of all samples seen so far. The complete procedure is detailed in Algorithm~\ref{alg:er}.

\subsection{ASER}
While ER provides a simple and effective foundation for replay-based continual learning, its reliance on uniform random sampling treats all stored exemplars as equally valuable. In practice, however, samples differ substantially in their contribution to preventing catastrophic forgetting: exemplars that are representative of their class distribution or that occupy critical regions near decision boundaries are far more informative during rehearsal than outliers or redundant samples.

Adversarial Shapley Value Experience Replay (ASER)~\cite{shim2021online} addresses this limitation by replacing both random memory retrieval and reservoir-based memory update with importance-driven selection based on the shapley values.

The Shapley value was originally developed in cooperative game theory.
Given a dataset of $N$ samples indexed by $\mathcal{I} = \{1, \ldots, N\}$ and a utility function $v(\mathcal{S})$ representing model performance when trained on subset $\mathcal{S} \subseteq \mathcal{I}$, the Shapley value for sample $i$ is defined as:
\begin{equation}
s(i) = \sum_{\mathcal{S} \subseteq \mathcal{I} \setminus \{i\}} \frac{|\mathcal{S}|!(N - |\mathcal{S}| - 1)!}{N!} \left[ v(\mathcal{S} \cup \{i\}) - v(\mathcal{S}) \right]
\end{equation}
This formulation computes the average marginal contribution of sample $i$ across all possible subsets that exclude it.
However, computing exact Shapley values is computationally prohibitive, requiring $\mathcal{O}(2^N)$ evaluations.

ASER addresses this by leveraging an efficient k-nearest neighbour (KNN) based approximation that reduces complexity to $\mathcal{O}(N \log N)$.
This approximation exploits the natural clustering of classes in the embedding space and defines a KNN utility function that measures the likelihood of correct classification. The resulting KNN Shapley value $s_j(i)$ quantifies the average marginal contribution of candidate sample $i$ with respect to a reference sample $j$.
When $i$ and $j$ share the same label then $s_j(i) > 0$, with magnitude $|s_j(i)|$ reflecting their proximity in the latent space.
Building on this, ASER introduces the Adversarial Shapley Value (ASV) score to balance the competing objectives of continual learning: maintaining stability on previously learned tasks while enabling
plasticity for new tasks. 

The ASV score is formulated as:
\nopagebreak
\begin{equation}
\label{eq:asv_1}
ASV(i) = \max_{j \in \mathcal{S}_{\text{sub}}} s_j(i) - \min_{k \in \mathcal{B}_n} s_k(i)
\end{equation}

where $i \in \mathcal{M} \setminus \mathcal{S}_{\text{sub}}$, $\mathcal{S}_{\text{sub}}$ is a class-balanced subsample from memory $\mathcal{M}$ and $\mathcal{B}_n$ is the incoming batch.
The first term captures stability while the second term captures plasticity.

We used the softer variant of ASV, which replaces the $\max$ and $\min$ operators with mean aggregation over the same sets, providing more stable estimates:
\begin{equation}
\label{eq:asv_2}
ASV_{\mu}(i) = \frac{1}{|\mathcal{S}_{\text{sub}}|} \sum_{j \in \mathcal{S}_{\text{sub}}} s_j(i) - \frac{1}{|\mathcal{B}_n|} \sum_{k \in \mathcal{B}_n} s_k(i).
\end{equation}

ASER modifies two critical components of the generic ER procedure described in Algorithm~\ref{alg:er}.
\textbf{1. Memory Retrieval.} Instead of random sampling, ASER computes ASV(i) as defined in Eq. \ref{eq:asv_2} (in Eq. \ref{eq:asv_1} at the original variant) for candidate samples in $\mathcal{M}$ and retrieves the $\mathcal{B}_m$ samples with the highest scores.
\textbf{2. Memory Update.} Rather than reservoir sampling, ASER employs a value maximization strategy.  It considers the union of the current memory and the incoming batch, $\mathcal{C} = \mathcal{M} \cup \mathcal{B}_n$, and computes the average KNN-SV score:
$v(i) = \frac{1}{|\mathcal{S}_{\text{sub}}|} \sum_{j \in \mathcal{S}_{\text{sub}}} s_j(i)$
for each candidate $i \in \mathcal{C}$. 
Here, the evaluation set $\mathcal{S}_{\text{sub}} \subset \mathcal{M}$ is a random subsample drawn from the current memory, serving as a reference for class representation. (We used the original process~\cite{shim2021online} for the size and construction rule of $\mathcal{S}_{\text{sub}}$.) The buffer is updated by retaining the top-$|\mathcal{M}|$ samples with the highest $v(i)$.
Together, these two modifications transform ER from a passive, random rehearsal mechanism into an active, value-driven memory management system, achieving state-of-the-art performance on the TSCIL benchmark \cite{qiao2024class, szHucs2025active}. 

\begin{algorithm}[H]
\caption{ACIL Framework for Task $t$}
\label{alg:acil}

\SetKwInput{Input}{Input}
\SetKwComment{Comment}{// }{}

\Input{Task $t$, unlabeled pool $\mathcal{D}_t$, consolidated memory $\mathcal{M}$, previous model $\theta_{t-1, N_a}$, AL policy $\mathcal{A}$, batch size $b$, cycles $N_a$}

\Comment{1. Initialization}
$\theta_{t,0} \leftarrow \theta_{t-1,N_a}$ \Comment*[r]{Parameters from end of task $t - 1$}
$L_{t,0} \leftarrow \emptyset$, $L^{\text{acc}}_{t,0} \leftarrow \emptyset$\;
$\mathcal{U}_{t,0} \leftarrow \mathcal{D}_t$ \Comment*[r]{Initialize unlabeled pool}

\Comment{2. Active learning cycles}
\For{$a = 1$ \KwTo $N_a$}{
    \Comment{2.1. Query and Update}
    $\mathcal{Q}_{t,a} \leftarrow \mathcal{A}(\mathcal{U}_{t,a-1}, \theta_{t,a-1}, b)$ \Comment*[r]{Select informative batch}
    Acquire labels for $\mathcal{Q}_{t,a}$ from oracle\;
    $L_{t,a} \leftarrow \{(x, y) \mid x \in \mathcal{Q}_{t,a}\}$ \Comment*[r]{Form current training batch}
    $L^{\text{acc}}_{t,a} \leftarrow L^{\text{acc}}_{t,a-1} \cup L_{t,a}$ \Comment*[r]{Append to cumulative set}
    $\mathcal{U}_{t,a} \leftarrow \mathcal{U}_{t,a-1} \setminus \mathcal{Q}_{t,a}$\; 

    \Comment{2.2. Incremental Training Step}
    $\theta' \leftarrow \theta_{t,a-1}$\;
    \While{not converged}{
        \For{$B_\ell \sim L^{\text{acc}}_{t,a}$}{
            \eIf{$t = 1$}{
                $\theta' \leftarrow \text{SGD}(B_\ell, \theta')$\;
            }{
                $B_m \leftarrow \text{MemoryRetrieval}(\mathcal{M}, B_\ell, \theta')$ \Comment*[r]{Retrieve replay batch}
                $\theta' \leftarrow \text{SGD}(B_m \cup B_\ell, \theta')$ \Comment*[r]{Train with replay}
            }
        }
    }
    $\theta_{t,a} \leftarrow \theta'$ \Comment*[r]{Update model state for cycle $a$}
}

\Comment{3. Memory Consolidation}
\For{$B_\ell \sim L^{\text{acc}}_{t,N_a}$}{
    $\mathcal{M} \leftarrow \text{MemoryUpdate}(\mathcal{M}, B_\ell, \theta')$ \Comment*[r]{Update consolidated memory}
}

\KwRet $\theta_{t,N_a}$, $\mathcal{M}$
\end{algorithm}

\subsection{ACIL Framework}
Standard continual learning methods, including both ER and ASER, operate under the assumption that all incoming data is fully labeled. While this assumption simplifies the learning protocol, it is rarely satisfied in practice. In real-world deployments, data acquisition is often automatic and inexpensive, sensors continuously generate readings, cameras capture frames, and devices log activity, but annotation requires domain expertise, sustained manual effort, and considerable time and financial resources. This disconnect between methodological assumptions and deployment constraints severely limits the practical applicability of existing CIL systems.

Active learning is the natural complement to this problem, rather than labeling everything, the learner strategically selects the most informative samples for annotation, maximizing learning performance under a fixed budget. 

We therefore propose integrating active learning directly into replay-based approaches, thereby introducing the Active Class Incremental Learning (ACIL) framework, as shown in Figure \ref{fig:scheme}.
At each task $t$, the learner has access to an unlabeled pool $\mathcal{D}_t$ of $N_t$ samples but may only query labels for at most $B \ll N_t$ of them.
In other words, we assume a limited annotation budget $B$ per task, which must be distributed among multiple query cycles.
The central challenge lies in designing query strategies that select samples that are simultaneously informative for learning the current task and representative enough to populate a memory buffer that supports long-term retention across all previously seen classes.

\vspace{\baselineskip}

\textbf{Active Learning Cycle}
Formally, we structure the annotation process within each task $t$ as a sequence of $N_a$ active learning cycles, as shown in Figure \ref{fig:scheme}. The dataset $\mathcal{D}_t$ is partitioned into $q$ equal slices, defining a query batch size of $b = \lfloor N_t / q \rfloor$. The learner is permitted to query $N_a \leq q$ of these slices, resulting in a total annotation budget of $B = N_a \times b$ and a budget ratio $\rho = N_a / q$ representing the fraction of the total data that is annotated.

\textbf{Query Batch $(Q_{t,a})$}: At cycle $a$ of task $t$, an active learning policy $\mathcal{A}$ selects a query set $Q_{t,a}$ of $b$ samples from the current unlabeled pool $\mathcal{U}_{t,a-1}$:
$$Q_{t,a} = \mathcal{A}(\mathcal{U}_{t,a-1},\, \theta_{t,a-1},\, b)$$
where $\theta_{t,a-1}$ denotes the model state at the end of the previous cycle. 

\textbf{Labeled Set} \; $(L_{t,a})$: Labels are then acquired from an oracle, representing a human annotator or equivalent labeling process, to form the current labeled batch:
$$L_{t,a} = \{(\mathbf{x}, y) \mid \mathbf{x} \in Q_{t,a}\}$$

\textbf{Cumulative Labeled Set} \; $(L^{\mathrm{acc}}_{t,a})$
This batch is appended to the cumulative labeled set, which aggregates all annotations acquired within the current task up to cycle $a$:
$$L^{acc}_{t,a} = L^{acc}_{t,a-1} \cup L_{t,a}$$
Crucially, model training at each cycle operates on the full cumulative set $L^{acc}_{t,a}$ rather than solely on the most recent batch $L_{t,a}$. This design choice ensures that the model retains knowledge of all classes encountered within the current task, regardless of when their samples were queried.

\vspace{\baselineskip}

Algorithm~\ref{alg:acil} presents the complete ACIL framework for a single task $t$. The procedure unfolds in three phases.
\\
\textbf{1. Initialization.} The model parameters $\theta_{t,0}$ are inherited from the end of the previous task, $\theta_{t-1,N_a}$. Both the current labeled batch $L_{t,0}$ and the cumulative set $L^{acc}_{t,0}$ are initialized as empty, and the full dataset $\mathcal{D}_t$ constitutes the initial unlabeled pool $\mathcal{U}_{t,0}$.
\\
\textbf{2. Cycle loop.} For each cycle $a \in \{1, \ldots, N_a\}$, the protocol proceeds as follows. First, the active learning policy $\mathcal{A}$ selects $b$ samples from the unlabeled pool, which are labeled by the oracle and appended to the cumulative set. The model then trains to convergence on $L^{acc}_{t,a}$: for the first task, each mini-batch $B_\ell$ is used directly for gradient updates; for subsequent tasks, every mini-batch is augmented with a replay batch $B_M$ retrieved from $\mathcal{M}$, and the model is updated on the combined batch $B_M \cup B_\ell$. The updated model state $\theta_{t,a}$ is then carried forward to the next cycle, where it serves as the basis for the next query.
\\
\textbf{3. Memory consolidation.} Once all $N_a$ cycles are complete, the consolidated memory $\mathcal{M}$ is updated by iterating through $L^{acc}_{t,N_a}$ in mini-batches using the strategy-specific \textit{Memory Update} procedure. The cumulative labeled set is then discarded, as its role is fulfilled by $\mathcal{M}$ for all future tasks.

Summarizing the Algorithm~\ref{alg:acil}, the model is updated after each active learning cycle, since new annotations are added to the system’s training process in each active learning cycle (not after all active learning cycles have been completed), as shown in lines 11-18. However, it is sufficient to update the consolidated replay memory after these cycles have been completed, but this does not cause a delay in the temporal integration of the acquired samples, since, as shown in line 17, the new and the memory instances together influence the training.

\vspace{\baselineskip}

\textbf{Memory Retrieval and Memory Update}
A critical design decision in the ACIL framework concerns how the memory $\mathcal{M}$ interacts with the active learning process. 
In our framework, \textit{MemoryRetrieval} and \textit{MemoryUpdate} procedures in Algorithm~\ref{alg:acil} are parts of the chosen  continual learning method method, so their implementation-level details were discussed at the ER and ASER methods.

During the cycle loop, \textit{Memory Retrieval} is invoked at every training iteration to augment the current mini-batch $B_{\ell}$ with a replay batch $B_M$ drawn from $\mathcal{M}$. Under ER, this retrieval is uniform random; under ASER, it is importance-driven, prioritizing exemplars with the highest ASV scores. 

\textit{Memory Update}, by contrast, is invoked only once per task, at the memory consolidation phase after all $N_a$ active learning cycles have been completed. At this point, the full cumulative labeled set $L^{\mathrm{acc}}_{t,N_a}$, comprising all samples annotated during the task, is streamed through the update procedure in mini-batches. Under ER, reservoir sampling ensures that a uniformly random subset of all seen samples is retained. Under ASER, a value-maximization strategy retains the top-$|\mathcal{M}|$ samples according to their KNN Shapley value.

\subsection{TypiCore}
From the experimental analysis of individual active learning strategies showed later in Section \ref{sec:results} emerges a fundamental limitation.
Specifically, no single method is sufficient on its own for the active class-incremental learning setting.

Uncertainty-based approaches systematically fail to outperform random selection.
Also, distribution-aware methods underperform with Core-Set that maximizes coverage, ensuring plasticity and immediate learning accuracy, but at the cost of catastrophic forgetting, as the memory becomes populated with atypical, boundary-adjacent samples that lack the robust features necessary for long-term retention.
On the contrary,  TypiClust, which builds a stable and representative memory buffer but underrepresents regions of the feature space.

Therefore, we propose a novel approach, TypiCore, that considers a trade-off of the two objectives: \textbf{representativeness} and \textbf{coverage}. \textit{Representativeness} prioritizes the selection of high-density, prototypical samples that accurately reflect the core structure of each class, ensuring stability as the model encounters new tasks. \textit{Coverage}, instead, focuses on selecting samples from underrepresented regions of the feature space, preserving diversity and preventing the memory buffer from collapsing onto a narrow, potentially unrepresentative subset of the data distribution. Neither objective alone is adequate: pure representativeness leads to a memory that is stable but lacks diversity, while pure coverage leads to diversity without the robustness needed to resist forgetting.

TypiCore addressed these two complementary distribution-aware objectives through a dynamic alternating mechanism that interleaves two phases across successive active learning cycles.
Concretely, the following two phases alternate within each task:
\begin{enumerate}
    \item A TypiClust selection step to find the most typical samples from the densest regions of the current unlabeled pool.
    \item A Core-Set selection step to identify samples that maximally extend the geometric coverage of the already-labeled set by applying the greedy k-center heuristic.
\end{enumerate}

This strict alternation prevents any prolonged period of strategic misalignment and ensures that the cumulative labeled set simultaneously captures the distributional core of each class and spans the broader feature space.

\vspace{\baselineskip}

The choice to initialize the alternation sequence with TypiClust rather than Core-Set is not arbitrary. At the onset of the first active learning cycle for each task, the model has had no opportunity to observe any labeled samples from the incoming classes, placing it in a cold-start condition where prediction-based methods are unreliable. In this regime, beginning with TypiClust ensures that the first queried samples are prototypical, representatives of the underlying distribution. 

This initialization provides a stable foundation for the subsequent Core-Set phase, which can then exploit the labeled anchors already placed in high-density regions to identify the largest remaining unlabeled voids and extend coverage outward. 

In summary, the TypiCore method determines only which unlabeled instances are queried and used to fine-tune the model within a task; thus, it has only an indirect effect on the consolidated memory. The final memory composition is determined by the ER or ASER's own selection rule, but both can only work from instances that have passed through the AL procedure, so TypiCore performs a kind of filtering task.

\begin{table}[htbp]
\centering
\caption{Overview of the TSCIL benchmark datasets. $d$ denotes the channel dimensionality and $\tau$ the sequence length.}
\label{tab:dataset_overview}
\begin{tabular}{lccccc}
\toprule
\textbf{Dataset} & \textbf{Shape ($d \times \tau$)} & \textbf{Train size} & \textbf{Test size} & \textbf{\#Classes} & \textbf{\#Tasks} \\
\midrule
UCI-HAR  & $9 \times 128$  & 5,952  & 2,947 & 6  & 3 \\
UWave    & $3 \times 315$  & 722    & 3,582 & 8  & 4 \\
GRABMyo  & $28 \times 128$ & 20,315 & 7,525 & 10 & 5 \\
WISDM    & $3 \times 200$  & 10,825 & 4,010 & 12 & 6 \\
\bottomrule
\end{tabular}
\end{table}

\section{Experimental Setting}
\label{sec:exp_setting}

\subsection{Benchmark Datasets}

We evaluate our approach on the benchmark datasets established in the TSCIL (Time Series Class-Incremental Learning) framework \cite{qiao2024class}. These datasets encompass a diverse range of time series classification tasks, varying in sequence length, channel dimensionality, application domain, and complexity. All datasets consist of fixed-length multivariate time series with a consistent shape $d \times \tau$, where $d$ represents the number of channels and $\tau$ denotes the temporal length of each sequence. Table~5.1 provides a comprehensive overview of the datasets, including their dimensionality, train/test splits, class counts, and task divisions for the incremental learning setting.

\textbf{UCI-HAR} (Human Activity Recognition Using Smartphones) \cite{anguita2013human} contains sensor data
from smartphones worn by 30 subjects performing six daily activities (walking, walking upstairs,
walking downstairs, sitting, standing, and laying). The dataset comprises 9-channel
accelerometer and gyroscope readings sampled at 50 Hz, with each sequence containing
128 time steps. This dataset is widely used for evaluating activity recognition systems and
represents a relatively balanced, moderate-complexity classification task.

\textbf{UWave} is a gesture recognition dataset collected from accelerometer sensors \cite{middlehurst2026multiverse}. It contains
3-channel time series of length 315, representing 8 different hand gestures performed
by multiple users. Despite having the smallest training set among the benchmarks (896
samples), it presents challenges due to inter-subject variability and the temporal complexity
of gesture patterns.

\textbf{GRABMyo} is a hand gesture recognition dataset based on surface electromyography
(sEMG) signals \cite{jiang2022gesture}. It contains 28-channel sEMG recordings of length 128, capturing 16
different hand and wrist gestures. With over 36,000 training samples, GRABMyo is the
largest dataset in our benchmark suite and presents unique challenges due to the noisy
nature of sEMG signals and the subtle differences between certain gesture classes.

\textbf{WISDM} (Wireless Sensor Data Mining) \cite{kwapisz2011activity} is an activity recognition dataset collected
from smartphone and smartwatch accelerometer sensors. It features 3-channel time series
of length 200, representing 18 different activities including various forms of walking, running,
and daily tasks. The dataset’s class imbalance and real-world collection conditions
make it particularly challenging for incremental learning scenarios.
We preserve the original train/test splits established in the TSCIL \cite{qiao2024class} to ensure fair comparison with existing methods. 
For the incremental learning setting, classes are randomly shuffled and partitioned into mutually exclusive subsets corresponding to sequential tasks, as described in Section 4.1. This partitioning simulates realistic scenarios where new classes arrive over
time and must be learned without access to previous task data.

\subsection{Evaluation Metrics}
We follow the standard TSCIL task construction protocol. For each dataset, classes are
randomly permuted and partitioned into tasks with mutually exclusive label spaces. To
ensure robustness against class ordering effects, all experiments are repeated with five
independent runs with distinct random class orderings.
Let $acc_{i,j}$ denote the classification accuracy on the test set of task j after completing the
training of task i (where $j \le i$). We report the following metrics, averaged over the five
runs with 95\% confidence intervals:
\noindent \textbf{Average Accuracy ($A_i$):} The average accuracy at the end of task $i$ is defined as:
\begin{equation}
A_i = \frac{1}{i} \sum_{j=1}^{i} \text{acc}_{i,j}
\end{equation}
The final average accuracy $ACC$ corresponds to $A_T$ and serves as the primary summary metric for overall performance.

\vspace{1em}
\noindent \textbf{Learning Accuracy:} To assess the model's ability to acquire new knowledge (plasticity), we calculate the learning accuracy $\text{acc}_{i,i}$ on the current task immediately after learning it and the average learning accuracy $A_{\text{cur}}$, which is the average of the task accuracies:
\begin{equation}
A_{\text{cur}} = \frac{1}{T} \sum_{i=1}^{T} \text{acc}_{i,i}
\end{equation}

\vspace{1em}
\noindent \textbf{Average Forgetting ($F_i$):} Forgetting measures the performance drop on previous tasks. For a specific task $j < i$, where $i > 1$, forgetting is defined as:
\begin{equation}
f_{i,j} = \max_{k < i} \, \text{acc}_{k,j} - \text{acc}_{i,j}
\end{equation}
The average forgetting after task $i$ is:
\begin{equation}
F_i = \frac{1}{i - 1} \sum_{j=1}^{i-1} f_{i,j}
\end{equation}
$FT$ ($F_T$) defines the Final Average Forgetting after all tasks have been learned.

\begin{figure}[!thbp]
    \centering
    \includegraphics[width=0.99\textwidth]{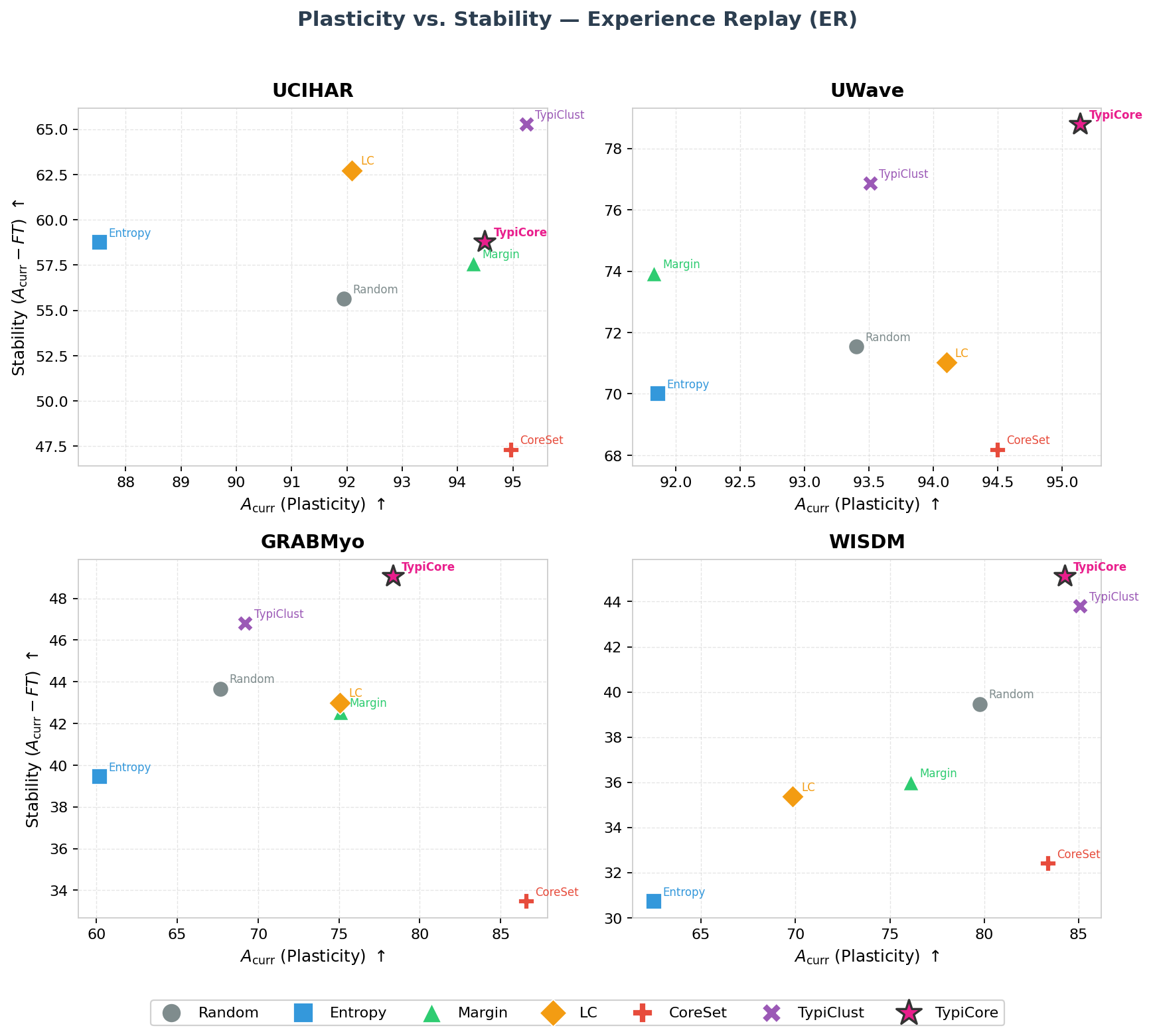}
    \caption{Plasticity vs.\ Stability trade-off for each dataset under the Experience Replay (ER) approach.  Methods closer to the top-right corner achieve a better balance between learning new tasks (plasticity) and retaining knowledge of previous ones (stability).} 
    \label{fig:plasticity_stability_er}
\end{figure}

Moreover, we define the \textbf{Stability} metric as the retention of previously acquired knowledge after learning all tasks:
\begin{equation}
    \text{Stability} = A_{\text{cur}} - FT
\end{equation}
where $A_{\text{cur}}$ is the average accuracy on the current task and $FT = F_T$ is the Final Average Forgetting, measuring the average performance drop on previous tasks after the full task sequence has been learned.
This metric is shown in Figure \ref{fig:plasticity_stability_er} to compare all the methods under the stability (Stability) and plasticity ($A_{cur}$). The results achieved on the individual components of Stability ($FT$ and $A_{\text{cur}}$) are presented in detail later in the Tables~\ref{tab:results_er} and~\ref{tab:results_aser}.

\subsection{Model Architecture}
We employ the 1D-CNN backbone from the TSCIL framework \cite{qiao2024class} across all experiments. The network $f_\theta$ is composed of a feature extractor $f_{\text{ext}} : \mathbb{R}^{d \times \tau} \to \mathbb{R}^h$ and a classifier $f_{\text{cl}} : \mathbb{R}^h \to \mathbb{R}^{|\mathcal{C}_{\le t}|}$, such that $f_\theta(x) = f_{\text{cl}}(f_{\text{ext}}(x))$.

The feature extractor $f_{\text{ext}}$ consists of four convolutional blocks. Each block contains a 1D convolutional layer (kernel size 5), Batch Normalization (BN), ReLU activation, 1D max-pooling, and dropout as presented in paper \cite{qiao2024class}. We used common embedding space for labeled and unlabeled instances, where the representations coming from the model without linear classification head. This model maps each input window to a 128-dimensional embedding, followed by a single linear classification head that is incrementally expanded with new output units as new task classes are introduced; the network is trained with cross-entropy loss. So, the embeddings obtained after pooling by omitting the head from the model used for classification. We did not use self-supervised representation learning, because at the beginning, the entire available data volume would have been needed (union of labeled and unlabeled instances), but in our case, the data only arrives per task. That is, the entire data set is not available at the beginning, the embedding is recalculated in every AL cycle, and becomes increasingly accurate as the model learns.

\subsection{Training Configuration}

We control the granularity of the active learning process via two parameters: the number of query cycles $N_a$ and the total number of dataset partitions $q$. The query batch size is derived as $b = \lceil \frac{N_t}{q} \rceil$, where $N_t$ is the size of the current task’s dataset. The optimization mini-batch size $B_\ell$ is set equal to the query batch size, i.e., $|B_\ell| = b$, and is sampled from the cumulative labeled set $\mathcal{L}_{t,a}^{\text{acc}}$ during training. For tasks $t > 1$, we maintain a balanced replay strategy with $|B_\ell| = |B_M|$, where $B_M$ denotes the memory replay batch, resulting in a total mini-batch size of $|B| = 2b$.

The configurations were chosen to balance annotation costs with learning stability. UCI-HAR employs fine-grained partitioning ($q = 300$) with a strict budget ratio ($\rho = 0.01$), whereas other datasets utilize coarser partitioning ($q \in \{20, 25, 40\}$) with ratios between $0.12$ and $0.30$. During the active learning cycles, the unlabeled pool gradually contracts, but in our experiments the budget ratio was only 1 percent, so this small shrinkage did not cause a relatively large change (experiments were finished after 5 cycles), i.e. the investigated methods were not affected by it.

We used 0.5 for $\lambda$ (balanced weights) and utilized the following optimization protocol:
\begin{itemize}
    \item \textbf{Optimizer:} Adam with an initial learning rate of $10^{-3}$.
    \item \textbf{Stopping Criterion:} Training proceeds for up to $100$ epochs per cycle, utilizing early stopping based on validation loss to prevent overfitting on small query batches.
\end{itemize}

The same backbone architecture, hyperparameters, and training schedule are used for all memory-based baselines (e.g., ER, ASER) to ensure fair comparison of query strategies.

\textbf{Memory Buffer Configuration}
Similar to the TSCIL benchmark,  we retain the original CIL replay-memory configuration, where the memory is nominally set to $5\%$ of the training-set size. While active learning imposes strict annotation budgets, this does not necessitate reducing the consolidated memory capacity.

\begin{table*}[ht]
\centering
\resizebox{\textwidth}{!}{%
\begin{tabular}{ll|ccccccc}
\toprule
\multirow{2}{*}{Dataset} & \multirow{2}{*}{Metric} & & \multicolumn{3}{c}{Uncertainty-based} & \multicolumn{2}{c}{Distribution-aware} & \\
\cmidrule(lr){4-6} \cmidrule(lr){7-8}
& & Random & Entropy & Margin & LC & CoreSet & TypiClust & \shortstack{TypiCore \\ (ours)} \\
\midrule
\multirow{3}{*}{UCIHAR}
&ACC$\uparrow$                & $67.84$ & $69.54$ & $69.83$ & $72.52$ & $63.19$ & $\mathbf{75.26}$ & $70.70$ \\
& FT $\downarrow$              & $36.30$ & $\mathbf{28.79}$ & $36.70$ & $29.36$ & $47.66$ & $29.97$ & $35.69$ \\
& $A_{\text{cur}}$ $\uparrow$ & $91.95$ & $87.53$ & $94.29$ & $92.09$ & $94.97$ & $\mathbf{95.24}$ & $94.49$ \\
\midrule
\multirow{3}{*}{UWave}
&ACC$\uparrow$                & $77.01$ & $75.46$ & $78.45$ & $76.79$ & $74.77$ & $81.03$ & $\mathbf{82.88}$ \\
& FT $\downarrow$              & $21.85$ & $21.86$ & $17.90$ & $23.08$ & $26.32$ & $16.64$ & $\mathbf{16.35}$ \\
& $A_{\text{cur}}$ $\uparrow$ & $93.40$ & $91.86$ & $91.83$ & $94.10$ & $94.50$ & $93.51$ & $\mathbf{95.14}$ \\
\midrule
\multirow{3}{*}{GRABMyo}
&ACC$\uparrow$                & $51.25$ & $44.87$ & $49.17$ & $49.58$ & $44.09$ & $54.89$ & $\mathbf{56.10}$ \\
& FT $\downarrow$              & $23.99$ & $\mathbf{20.76}$ & $32.54$ & $32.06$ & $53.11$ & $22.40$ & $29.27$ \\
& $A_{\text{cur}}$ $\uparrow$ & $67.66$ & $60.21$ & $75.09$ & $75.06$ & $\mathbf{86.57}$ & $69.20$ & $78.34$ \\
\midrule
\multirow{3}{*}{WISDM}
&ACC$\uparrow$                & $46.41$ & $36.02$ & $42.66$ & $41.13$ & $40.90$ & $50.69$ & $\mathbf{51.73}$ \\
& FT $\downarrow$              & $40.29$ & $\mathbf{31.78}$ & $40.12$ & $34.47$ & $50.99$ & $41.27$ & $39.12$ \\
& $A_{\text{cur}}$ $\uparrow$ & $79.76$ & $62.50$ & $76.10$ & $69.85$ & $83.39$ & $\mathbf{85.08}$ & $84.25$ \\
\bottomrule
\end{tabular}%
}
\caption{Results across datasets and query strategies for Experience Replay (ER)}
\label{tab:results_er}
\end{table*}

\begin{table*}[ht]
\centering
\resizebox{\textwidth}{!}{%
\begin{tabular}{ll|ccccccc}
\toprule
\multirow{2}{*}{Dataset} & \multirow{2}{*}{Metric} & & \multicolumn{3}{c}{Uncertainty-based} & \multicolumn{2}{c}{Distribution-aware} & \\
\cmidrule(lr){4-6} \cmidrule(lr){7-8}
& & Random & Entropy & Margin & LC & CoreSet & TypiClust & \shortstack{TypiCore \\ (ours)} \\
\midrule
\multirow{3}{*}{UCIHAR}
&ACC$\uparrow$                & $63.00$ & $61.10$ & $62.57$ & $57.46$ & $63.03$ & $62.52$ & $\mathbf{69.11}$ \\
& FT $\downarrow$              & $46.24$ & $43.66$ & $45.80$ & $\mathbf{53.15}$ & $46.99$ & $47.02$ & $38.89$ \\
& $A_{\text{cur}}$ $\uparrow$ & $93.83$ & $88.50$ & $92.87$ & $92.90$ & $94.36$ & $93.86$ & $\mathbf{95.04}$ \\
\midrule
\multirow{3}{*}{UWave}
&ACC$\uparrow$                & $59.75$ & $60.96$ & $61.40$ & $62.53$ & $61.10$ & $62.26$ & $\mathbf{63.65}$ \\
& FT $\downarrow$              & $43.23$ & $41.85$ & $44.29$ & $\mathbf{40.00}$ & $44.67$ & $41.17$ & $42.39$ \\
& $A_{\text{cur}}$ $\uparrow$ & $92.18$ & $92.35$ & $94.62$ & $92.53$ & $94.60$ & $93.14$ & $\mathbf{95.44}$ \\
\midrule
\multirow{3}{*}{GRABMyo}
&ACC$\uparrow$                & $49.79$ & $42.83$ & $46.07$ & $46.06$ & $41.19$ & $\mathbf{52.91}$ & $52.84$ \\
& FT $\downarrow$              & $31.38$ & $\mathbf{23.31}$ & $27.54$ & $30.42$ & $53.85$ & $29.80$ & $32.88$ \\
& $A_{\text{cur}}$ $\uparrow$ & $73.38$ & $58.82$ & $67.26$ & $69.68$ & $\mathbf{84.18}$ & $76.35$ & $78.18$ \\
\midrule
\multirow{3}{*}{WISDM}
&ACC$\uparrow$                & $42.93$ & $37.56$ & $42.30$ & $39.39$ & $39.65$ & $47.37$ & $\mathbf{47.98}$ \\
& FT $\downarrow$              & $42.88$ & $\mathbf{28.34}$ & $40.40$ & $33.75$ & $51.31$ & $44.00$ & $42.35$ \\
& $A_{\text{cur}}$ $\uparrow$ & $78.37$ & $60.07$ & $75.97$ & $67.51$ & $82.41$ & $\mathbf{84.04}$ & $83.27$ \\
\bottomrule
\end{tabular}%
}
\caption{Results across datasets and query strategies for ASER}
\label{tab:results_aser}
\end{table*}

\section{Results}
\label{sec:results}

\begin{table*}[htbp]
\centering
\begin{tabular}{lcccccc}
\toprule
 & \multicolumn{2}{c}{\textbf{ER}} & \multicolumn{2}{c}{\textbf{ASER}} & \multicolumn{2}{c}{\textbf{\shortstack{ER \& \\ ASER}}} \\
\cmidrule(lr){2-3} \cmidrule(lr){4-5} \cmidrule(lr){6-7}
\textbf{Method} & \textbf{ACC} & \textbf{$A_{\text{cur}}$} & \textbf{ACC} & \textbf{$A_{\text{cur}}$} & \textbf{ACC} & \textbf{$A_{\text{cur}}$} \\
\midrule
\textbf{Random} & $4.00$ & $5.25$ & $4.00$ & $4.75$ & $4.00$ & $5.00$ \\
\textbf{Entropy} & $6.00$ & $6.75$ & $6.25$ & $6.75$ & $6.13$ & $6.75$ \\
\textbf{Margin} & $4.00$ & $4.75$ & $4.00$ & $4.75$ & $4.00$ & $4.75$ \\
\textbf{LC} & $4.00$ & $4.50$ & $5.00$ & $5.25$ & $4.50$ & $4.88$ \\
\textbf{CoreSet} & $6.75$ & $\mathbf{2.00}$ & $4.75$ & $2.25$ & $5.75$ & $2.13$ \\
\textbf{TypiClust} & $1.75$ & $2.75$ & $2.75$ & $2.75$ & $2.25$ & $2.75$ \\
\textbf{\shortstack{TypiCore \\ (ours)}} & $\mathbf{1.50}$ & $\mathbf{2.00}$ & $\mathbf{1.25}$ & $\mathbf{1.50}$ & $\mathbf{1.38}$ & $\mathbf{1.75}$ \\
\bottomrule
\end{tabular}
\caption{Comparison of ranks of methods (averages on all datasets) with ER and ASER variants, and union of variants. Smaller is better, and the best average rank in each column is highlighted by bold characters.}
\label{tab:results_method_ranks}
\end{table*}

\begin{table*}[ht]
\centering
\scriptsize
\resizebox{\textwidth}{!}{%
\begin{tabular}{l|cccc}
\toprule
Method & UCIHAR & UWave & GRABMyo & WISDM \\
\midrule

\textit{Fine-Tuning}   & \textit{32.9} & \textit{26.0} & \textit{19.4} & \textit{15.5}  \\
\textit{Joint Training}  & \textit{93.9} & \textit{96.6} & \textit{93.8} & \textit{85.7} \\

\midrule
\textit{LwF}     & \textit{40.0} & \textit{47.3} & \textit{19.4} & \textit{15.9} \\
\textit{MAS}     & \textit{48.0} & \textit{51.9} & \textit{16.9} & \textit{11.2} \\

\midrule
\textit{ER}   & \textit{72.8} & \textit{72.7} & \textit{46.5} & \textit{41.7} \\
\textit{ASER} & \textit{89.1} & \textit{83.2} & \textit{55.9} & \textit{51.6} \\

\midrule
Random (ER)   & 67.84 & 77.01 & 51.25 & 46.41 \\
Random (ASER) & 63.00 & 59.75 & 49.79 & 42.93 \\

\midrule
Unc-Entropy (*)    & 69.54 & 75.46 & 44.87 & 36.02 \\
Unc-Margin (*)     & 69.83 & 78.45 & 49.17 & 42.66 \\
Unc-LC (*)         & 72.52 & 76.79 & 49.58 & 41.13 \\

\midrule
CoreSet (*)    & 63.19 & 74.77 & 44.09 & 40.90 \\
TypiClust (*)  & \textbf{75.26} & 81.03 & 54.89 & 50.69 \\
TypiCore (ER)  & 70.70 & \textbf{82.88} & \textbf{56.10} & \textbf{51.73} \\
TypiCore (ASER) & 69.11 & 63.65 & 52.84 & 47.98 \\

\bottomrule
\end{tabular}%
}
\caption{ACC results across datasets including baselines, continual learning methods, and active learning strategies. TypiCore is reported under ER and ASER settings. The sign (*) means the better version among ER and ASER variants. CIL Reference corresponds to the active-task (AT) performance of the continual learning reference model under ER and ASER.}
\label{tab:at_full_transposed_final}
\end{table*}

\subsection{Uncertain-based Methods}
Tables~\ref{tab:results_er} and~\ref{tab:results_aser} report the full performance comparison across all query strategies under the ER and ASER replay settings, respectively. In each experiment, the annotation was limited according to the active learning scenario, which was solved differently: the Random method used random selection; the Uncertainty-based and Distribution-aware methods solved the problem based on scores specified in their algorithms. Across both ER and ASER, uncertainty-based strategies consistently fail to improve over random selection and frequently degrade performance.

Under ER, entropy sampling reduces ACC on GRABMyo from 51.25 to 44.87 and on WISDM from 46.41 to 36.02, a drop of more than ten percentage points on the latter.
Learning accuracy ($A_\text{cur}$) follows the same pattern: entropy with ER reaches only 62.50 on WISDM against the random baseline of 79.76. The degradation is equally pronounced under ASER. 

Margin sampling is the most competitive of the three uncertainty variants, occasionally matching or slightly exceeding the random baseline in ACC (e.g., 78.45 vs. 77.01 on UWave with ER), but it never establishes a consistent advantage.
Therefore, the systematic underperformance of uncertainty methods across datasets and replay strategies confirms that boundary-focused selection is fundamentally misaligned with the requirements of rehearsal-based continual learning.

\subsection{Distribution-aware Methods}
The results for distribution-aware methods are reported in Tables \ref{tab:results_er} and \ref{tab:results_aser} for ER and ASER, respectively, and the accuracy trends across sequential tasks on GRABMyo dataset are shown in Figure \ref{fig:plot_over_time_grabmyo} (all other datasets can be seen in the Appendix \ref{sec:8__Appendix}: UWave in Fig. A1, HAR in Fig. A2, and WISDM in Fig. A3).

\begin{figure}[!thbp]
    \centering
    \includegraphics[width=0.99\textwidth]{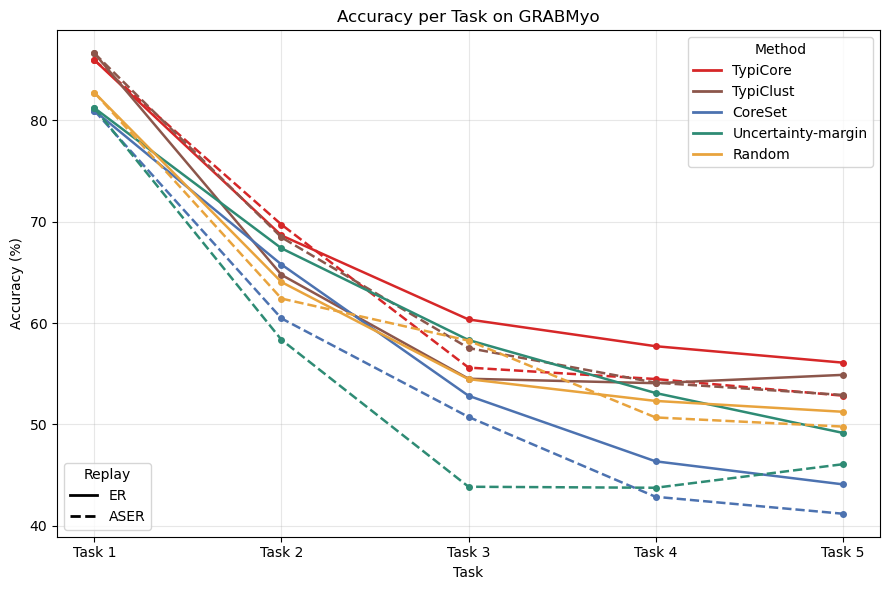}
    \caption{Average accuracy across sequential tasks on the GRABMyo dataset for four active learning query strategies (TypiClust, CoreSet, Uncertainty-margin, Random) combined with two replay mechanisms (ER: solid lines; ASER: dashed lines).TypiClust-ER achieves the highest final accuracy (55\%), while uncertainty-margin and CoreSet paired with ASER exhibit the steepest forgetting curves. }
    \label{fig:plot_over_time_grabmyo}
\end{figure}

CoreSet produces a sharp and consistent plasticity–stability trade-off. Under ER, it achieves the highest or near-highest $A_\text{cur}$ on several datasets, reaching 94.97 on UCI-HAR, 94.50 on UWave, and a striking 86.57 on GRABMyo, outperforming the random baseline by nearly nineteen percentage points on the latter. However, this immediate learning gain comes at a severe cost to long-term retention: FT on GRABMyo rises to 53.11 under ER and 53.85 under ASER, more than doubling the random baseline values of 23.99 and 31.38, respectively.

TypiClust presents the inverse profile. It consistently achieves competitive or best-in-class ACC values under ER, reaching 75.26 on UCI-HAR, 81.03 on UWave, 54.89 on GRABMyo, and 50.69 on WISDM, all above their respective random baselines. Its FT figures are also among the lowest across all strategies, particularly on UCI-HAR (29.97) and UWave (16.64).
 Under ASER, TypiClust's advantage over the random baseline largely disappears, with ACC values close to or below random on UCI-HAR (62.52 vs. 63.00) and UWave (62.26 vs. 59.75), suggesting that its benefits are more strongly realized when paired with the simpler replay mechanism.

\subsection{Our Approach}
Tables~\ref{tab:results_er} and~\ref{tab:results_aser} present that TypiCore achieves the best overall ACC on three out of four datasets under ER (82.88 on UWave, 56.10 on GRABMyo, 51.73 on WISDM) and under ASER as well, including a clear margin over all other methods on UCI-HAR (69.11 vs. the next best of 63.03 for CoreSet-ASER).

For cross-dataset generalization, we created a ranking based on the accuracy values of the Tables~\ref{tab:results_er} and \ref{tab:results_aser} by sorting all seven methods for each dataset (in the case of both accuracy metrics, i.e. ACC and $A_{cur}$, the method with the highest value received rank 1). The average rank of the indicators calculated on the four data sets is shown in the Table~\ref{tab:results_method_ranks}. We calculated these values for ER (from Table~\ref{tab:results_er}) and ASER (from Table~\ref{tab:results_aser}) variants of the methods, and for the union of them as can be seen in the different columns of Table~\ref{tab:results_method_ranks}. The best average rank in each column is highlighted by bold characters. The table shows that TypiCore achieved the best values among all competitors.

Figure~\ref{fig:plasticity_stability_er} further contextualizes these results by examining the 
plasticity--stability trade-off across all query strategies. TypiCore consistently occupies the 
top-right region of the plot, indicating that its accuracy gains do not come at the expense of 
catastrophic forgetting. Notably, CoreSet, despite achieving competitive $A_{\text{cur}}$ scores, 
suffers from substantially higher forgetting, resulting in lower stability across all four datasets.

\vspace{\baselineskip}
\subsection{Comparison among CL methods}
Table~\ref{tab:at_full_transposed_final} places the active learning results within the wider context of continual learning methods, where TypiCore-ER is the best on three datasets, and the second best on UCIHAR dataset. The methods in the first two blocks (marked in italics) are those that used full labeled dataset, unlike the methods below them, which only had a limited number of such instances available. The second block contains methods that were developed exclusively for continuous learning. Regularization-based approaches such as LwF \cite{li2017learning} and MAS \cite{aljundi2018memory} remain far below all rehearsal methods, with ACC values as low as 11.2 (MAS on WISDM) and 16.9 (MAS on GRABMyo).
Fine-tuning, as expected, collapses to near-chance performance across all datasets. Joint training (where there is no annotation limit and no forgetting, as the model learns all classes simultaneously), establishes the practical upper bound, reaching 93.9 on UCI-HAR and 93.8 on GRABMyo. 

Both TypiCore versions (ER and ASER) surpass the fully supervised LwF and MAS methods on all datasets, demonstrating that intelligent sample selection can compensate for the absence of full label supervision. One possible reason is that by systematically selecting prototypical and geometrically diverse samples, TypiCore effectively filters out noisy, redundant, or ambiguous annotations that, when included indiscriminately under full supervision.

A consistent finding across all strategies is that ER outperforms ASER in the active learning setting. The gap is particularly pronounced for TypiCore on UWave, where TypiCore-ER achieves 82.88 versus 63.65 for TypiCore-ASER. Under severe annotation budgets, the simpler and more robust random rehearsal mechanism of ER appears better suited to the constraints of the active learning regime.

\section{Conclusion}
\label{sec:conclusion}

\subsection{Summary of Contributions}

In this paper, we addressed the problem of Active Class-Incremental Learning (ACIL) for multivariate time series classification, a setting that combines the sequential nature of continual learning with the practical constraint of limited label availability. We conducted a systematic evaluation of a broad range of active learning query strategies integrated into rehearsal-based continual learning approaches.

We assessed their impact on plasticity, stability, and label efficiency across four benchmark datasets from the TSCIL framework.
Our analysis revealed several key insights. Uncertainty-based methods consistently fail to improve over random selection and frequently degrade performance in the ACIL setting, as boundary-focused sample selection is fundamentally misaligned with the requirements of memory-based replay. 
Distribution-aware methods have better performance, but neither strategy alone achieves a robust and consistent balance across datasets and task sequences.

Motivated by these findings, we proposed TypiCore, a novel hybrid query strategy that dynamically alternates between typicality-based and diversity-based selection across active learning cycles. By initializing each task with a TypiClust step to anchor prototypical samples in high-density regions, and subsequently applying CoreSet to extend coverage into underrepresented areas of the feature space, TypiCore constructs memory buffers that are simultaneously representative and diverse. 

Evaluated on the TSCIL benchmark, TypiCore achieves statistically significant improvements over all baselines under both ER and ASER replay settings, and on several datasets exceeds the performance of fully supervised continual learning methods while annotating a fraction of the available data. This shows that intelligent label selection can act as an implicit regularizer, filtering out noisy and redundant samples that would otherwise destabilize rehearsal-based learning.

Moreover, a consistent finding across all experiments is that the simpler ER mechanism outperforms ASER in the active learning regime, particularly under severe annotation budgets. This suggests that the importance-driven memory management of ASER, while effective under full supervision, may be less robust when the labeled pool is small and potentially unrepresentative in early cycles.

\subsection{Limitations and Future Directions}

Although the proposed method does not incorporate a data-dependent switching mechanism between TypiClust and Core-Set, future advancements will focus on developing such a rule to achieve better adaptability. One of the potential directions for future work is to develop such a rule, and another direction is the investigation of multichannel structure in time-series instances. The third direction is the aggravating circumstance with imbalance data.

Our current experiments were conducted on relatively balanced datasets. Class imbalance
presents a significantly different challenge where sampling mechanism of ASER could
demonstrate substantial advantages. In imbalanced scenarios, random selection of ER risks over-representing majority classes while under-sampling minority classes, leading to biased
models that perform poorly on rare but important categories. Balancing mechanisms of ASER
could prove particularly valuable when combined with active learning strategies that are aware of class distribution. Future work could investigate how different active learning strategies interact with ASER under varying degrees of class imbalance.

Task order robustness can also be examined in the future by testing the method in a variable task order environment with controlled difficulty levels to assess its resilience against the order effects commonly observed in continual learning settings.

Future work will explore additional promising directions. First, future methods could consider dynamic budget allocation: rather than distributing the annotation budget uniformly across tasks, methods could estimate the degree of distribution shift between incoming and previously seen data, and allocate more labeling resources to tasks that deviate most significantly from prior knowledge, as these are likely to require more annotated samples to be learned effectively without destabilizing the existing memory buffer.

Second, future work could investigate more complex scenarios that combine elements of both Domain-Incremental and Class-Incremental Learning, where the same classes may reappear across tasks with shifted distributions alongside the introduction of entirely new ones. In this setting, the model must simultaneously distinguish between genuine new knowledge and distributional variations of already-known classes, posing new challenges for both query strategy design and memory management.

\backmatter





\section*{Declarations}

\bmhead{Ethical Approval}
Not Applicable.

\bmhead{Data availability}
All the datasets mentioned in the manuscript for experimentation are publicly available. No new datasets were created in this research.

\bmhead{Materials availability}
The source code of our TypiCore method with the complete framework is publicly available on the following repository for reproducibility: https://anonymous.4open.science/r/TSCIL-AL-3D67

\bmhead{Competing interests}
The authors declare that they have no conflict of interest.

\bmhead{Authors' contributions}
G.Sz.: Conceptualization, Formal analysis, Investigation, Methodology, Writing – original draft, Writing – review \& editing, Validation, Supervision. S.J.: Methodology, Software, Writing – original draft, Validation. M.N.: Validation, Investigation, Supervision. D.D.P.: Validation, Visualization, Writing – review \& editing. G.A.S.: Project Administration, Validation. All authors wrote and reviewed the manuscript.




\bmhead{Acknowledgements}
Project No. KDP-IKT-2023-900-I1-00000957/0000003 has been implemented with the support provided by the Ministry of Culture and Innovation of Hungary from the National Research, Development and Innovation Fund, financed under the C2299763 funding scheme.



\bibliography{sn-bibliography}

\newpage
\begin{appendices}

\section{}
\label{sec:8__Appendix}

\begin{figure}[!thbp]
    \centering
    \includegraphics[width=0.99\textwidth]{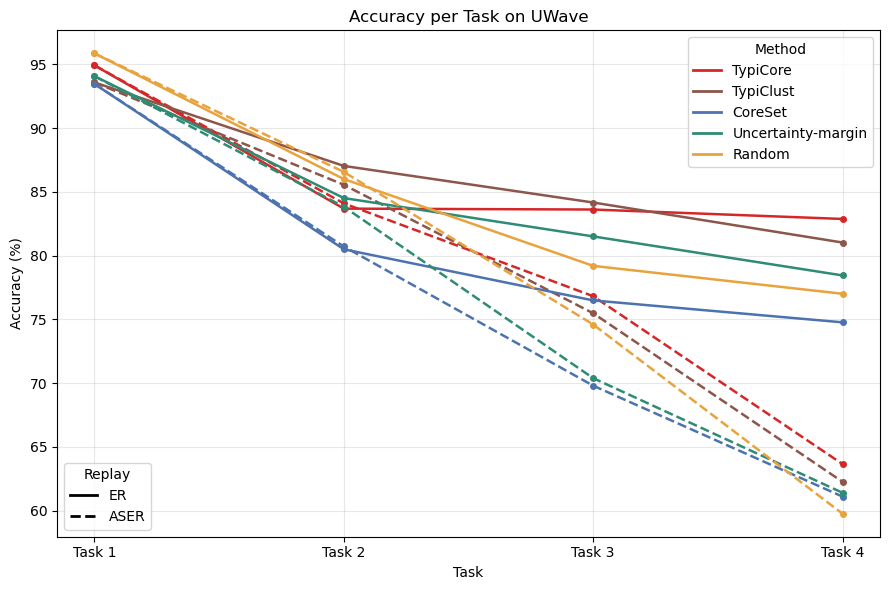}
    \caption{Average accuracy across sequential tasks on the UWave dataset for four active learning query strategies (TypiClust, CoreSet, Uncertainty-margin, Random) combined with two replay mechanisms (ER: solid lines; ASER: dashed lines).}
    \label{fig:plot_over_time_uwave}
\end{figure}

\begin{figure}[!thbp]
    \centering
    \includegraphics[width=0.99\textwidth]{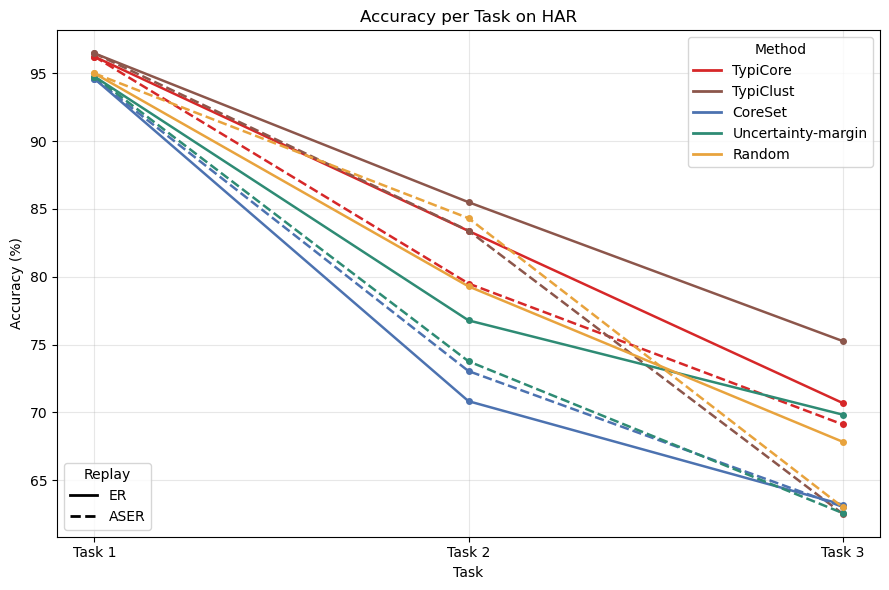}
    \caption{Average accuracy across sequential tasks on the HAR dataset for four active learning query strategies (TypiClust, CoreSet, Uncertainty-margin, Random) combined with two replay mechanisms (ER: solid lines; ASER: dashed lines).}
    \label{fig:plot_over_time_har}
\end{figure}

\begin{figure}[!thbp]
    \phantomsection
    \centering
    \includegraphics[width=0.99\textwidth]{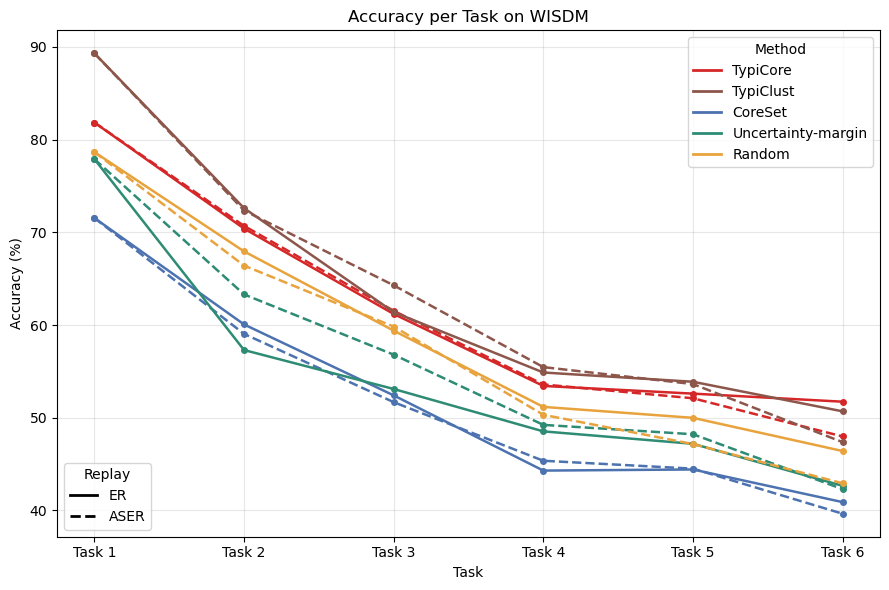}
    \caption{Average accuracy across sequential tasks on the WISDM dataset for four active learning query strategies (TypiClust, CoreSet, Uncertainty-margin, Random) combined with two replay mechanisms (ER: solid lines; ASER: dashed lines).}
    \label{fig:plot_over_time_wisdm}
\end{figure}


\end{appendices}

\end{document}